\documentclass[11pt]{article}
\usepackage{coling2018}
\usepackage{times}
\usepackage{url}
\usepackage{latexsym}
\usepackage{csquotes}
\usepackage{makecell}
\usepackage{graphicx}
\usepackage{floatrow}
\usepackage{caption}
\usepackage{comment}
\usepackage[font=small,labelfont=bf]{caption}

\newfloatcommand{capbtabbox}{table}[][\FBwidth]

\usepackage{blindtext}

\title{Who is Killed by Police: Introducing Supervised Attention for Hierarchical LSTMs}

\author{
Minh Nguyen$^\dag$ and Thien Huu Nguyen$^\#$$^\ddagger$ \\
 $^\dag$ Hanoi University of Science and Technology, Hanoi, Vietnam\\
 $^\#$ Montreal Institute for Learning Algorithms, University of Montreal, Canada \\
 $^\ddagger$ Department of Computer and Information Science, University of Oregon, USA \\
 {\small \tt 
minh.nv142950@sis.hust.edu.vn,thien@cs.uoregon.edu} }

\date{}

\begin{document}
\maketitle
\begin{abstract}
Finding names of people killed by police has become increasingly important as police shootings get more and more public attention (police killing detection). Unfortunately, there has been not much work in the literature addressing this problem. The early work in this field \cite{keith2017identifying} proposed a distant supervision framework based on Expectation Maximization (EM) to deal with the multiple appearances of the names in documents. However, such EM-based framework cannot take full advantages of deep learning models, necessitating the use of hand-designed features to improve the detection performance. In this work, we present a novel deep learning method to solve the problem of police killing recognition. The proposed method relies on hierarchical LSTMs to model the multiple sentences that contain the person names of interests, and introduce supervised attention mechanisms based on semantical word lists and dependency trees to upweight the important contextual words. Our experiments demonstrate the benefits of the proposed model and yield the state-of-the-art performance for police killing detection.
\end{abstract}

\section{Introduction}

\blfootnote{
    %
    %
    %
    %
    %
    %
    \hspace{-0.65cm}  
    This work is licensed under a Creative Commons 
    Attribution 4.0 International License.
    License details:
    \url{http://creativecommons.org/licenses/by/4.0/}
}




We study the problem of police killing detection from text. The key challenge is to be able to take a person name in the pool of documents (corpus) and automatically decide whether the corresponding person is killed by police or not based on the textual evidences in the corpus. For instance, the following sentence describes the police-caused death of ``\textit{Micah Jester}'': ``\textit{Old \textbf{Micah Jester} was fatally shot by APD officers.}''. This problem has drawn much public attention recently; however, it has not been investigated adequately by the natural language processing (NLP) community. To our knowledge, the only NLP work for police killing recognition so far is by \cite{keith2017identifying} who rely on machine learning models to perform the automatic detection.

It is challenging to apply the machine learning models in this case as identifying police killings from text is a relatively new problem in machine learning research with no available training datasets to supervise the models. The only sources of supervision on which we can rely for this problem are the current databases that record the names of the police-killed victims in the past. Among these databases, the Fatal Encounters\footnote{This database has been produced by D. Brian Burghart and colleagues by manually reading millions of news headlines and ledes.} (FE) database has emerged as the most comprehensive database with a relatively large number of recorded victims (over 23,000 victims). In order to take advantage of this database, \cite{keith2017identifying} employs distant supervision \cite{craven1999constructing,mintz2009distant} that extracts person names from a corpus and aligns them with the victim names in the database. The matched names are considered as corresponding to people killed by police (positive entities) while the non-matched names constitute the negative examples in a binary classification problem for names in police killing detection. As the name itself does not carry much information, each extracted name is associated with the set of sentences in which the name appears in the corpus. This set of sentences is called the sentence container for the corresponding name (person). The sentence containers along with the distant supervision labels (i.e, positive or negative) of the corresponding names would serve as the training data for the binary name classification problem.

An important characteristic of the sentence containers is that they might contain multiple sentences corresponding to the multiple appearances of the names in the corpus. Although all of these sentences mention the names of interests, some of them might not express police killing incidents. Consequently, it is crucial for the systems to be able to model the multiple sentences in the containers appropriately so that the correct sentences for police killing events can be captured to perform classification for the sentence containers of names. \cite{keith2017identifying} solves this problem by introducing a latent variable for each sentence in the container of a name to predict whether the sentence describes the person as having been killed by police or not. Such latent variables are then modeled by sentence level classifiers (i.e, logistic regressions and convolutional neural networks) and aggregated for the final prediction for the name. \cite{keith2017identifying} learns the parameters for the sentence classifiers via an Expectation Maximization (EM) based framework that alternates between estimating the latent variables and updating the model parameters. In \cite{keith2017identifying}, the authors demonstrate that the EM-based framework works well when the sentence level classifiers involve logistic regression with hand-crafted features. However, when deep learning models (i.e, convolutional neural networks) are employed for the sentence level classifiers, the performance of the EM-based framework drops significantly. We attribute this problem to the limitation of the EM-based framework to train the non-convex classifiers of deep learning, causing the inability to exploit the automatically learnt representations from deep learning and necessitating the use of complicated and laborious feature engineering.

In order to overcome this problem, we propose a novel deep learning framework for the problem of police killing recognition via hierarchical long short-term memory networks (LSTM) \cite{Hochreiter:97,yang2016hierarchical}. Our model does not make individual predictions on each sentence with the latent variables but involves a direct prediction to the name of interest based on its sentence container. Two layers of LSTMs are applied to model the sentence containers. The first LSTM layer learns the representations for the sentences in the containers by recurring over their words (the word level). The second LSTM layer, on the other hand, consumes the sentence representations (the sentence level) to produce container representations for police killing predictions. Attention mechanisms are then introduced into both LSTM layers to appropriately quantify the contribution of each word and sentence in the containers for the final predictions. This approach facilitates the modeling of multiple sentences in the containers, leading to the effective training of the deep learning models in a single framework and alleviating the reliance on manually designed features for this problem.


In the previous hierarchical LSTM models, attention scores are computed and normalized using the hidden representations of words and sentences generated by LSTMs \cite{yang2016hierarchical}. Unfortunately, for our problem of police killing recognition, this traditional attention computation tends to assign very high weight to the words in the names of interests and relatively low weights to the other important context words in the sentences. Such failure to adequately capture those context words would potentially lead to incorrect predictions for the containers. This problem is stem from the use of the position embeddings to specify the names of interests that might put too much emphasis on the current names. In order to solve this problem, we propose to integrate the supervised attention mechanisms into the hierarchical LSTM model that help to bias the attention scores toward the heuristically important words in the sentences (supervised attention) \cite{mi2016supervised}. In particular, we rely on linguistic intuitions to heuristically select the informative context words for the problem of police killing detection. These words are then used to guide the attention computation via penalizing the model parameters that generate low attention scores for such guidance words. We investigate several heuristics to choose the guidance words based on semantical word lists and dependency trees. The experiments show that the supervised attention mechanism with those heuristics helps to improve the performance of the hierarchical LSTM model and yield the state-of-the-art performance for the problem of detecting police killings. To the best of our knowledge, this is the first work that introduces supervised attention into the hierarchical LSTM models and employs semantical word lists and dependency trees to select guidance words.

\section{Related Work}
Although police killing recognition is a new task, it has some elements with the information extraction (IE) research of NLP that can be used to solve the task with some modifications. The most related IE task for police killing detection is event extraction that aims to detect events (i.e, marriage, attack, die etc.) in text \cite{li2014incremental,Nguyen:15b,Chen:15,Nguyen:16b}. Killings is one of the event types that the current event extraction systems can identify \cite{das2014frame,li2014incremental,Nguyen:16c}, allowing the detection of police killing incidents when appropriate adaptations are introduced. Unfortunately, such adaptations result in poor performance for police killing recognition as shown in \cite{keith2017identifying}.

Distant supervision is another element of IE that is employed in this research to generate training data for police killing detection. In particular, distant supervision has been used to produce training data for relation extraction \cite{craven1999constructing,bunescu2007learning,mintz2009distant,riedel2010modeling,Mihai:12} and event extraction \cite{reschke2014event}.


Hierarchical deep learning techniques that model both the word and sentence levels have been employed for several NLP tasks, including relation extraction \cite{Lin:16}, question answering \cite{Choi:17} and extractive summarization \cite{Cheng:16}. Such work often uses convolutional neural networks to operate at the work level. This is different from our proposed model for police killing detection that employs LSTMs and supervised attentions to acquire sentence representations for police killing recognition. Perhaps the most related model to ours is \cite{yang2016hierarchical} that utilizes hierarchical LSTMs for text categorization. Our model also relies on hierarchical LSTMs, but it is designed for police killing detection, characterizing position embeddings and supervised attentions to inject external knowledge (i.e, the heuristics for guidance words).

Finally, supervised attention mechanisms have been used recently for several natural language tasks. For machine translation, the attention guidance is based on word alignment \cite{mi2016supervised,liu2016neural} while entity mentions are chosen as the guidance words for event detection \cite{liu2017exploiting}. Our work in this paper is different as we consider supervised attention for police killing recognition using semantical word lists and dependency parsing trees \cite{schuster2016enhanced} to guide the attention components. Our model features hierarchical LSTMs to tackle distant supervision data that does not emerge in such prior work.

\section{Model}
We formalize the problem of finding people killed by police as follows.

Let $D$ be a set of documents (corpus), $E=\left\{e_{i}\right\}_{i=1}^{N}$ be the set of entities (people) whose names appear in $D$ ($N$ is the number of the entities in $E$), and $C=\left\{c_{i}\right\}_{i=1}^{N}$ be the set of sentence containers for the entities in $E$ (i.e, $c_i$ is the set of sentences that contain the name of the entity $e_i$ in $D$).

Each entity $e_i$ in $E$ has a label $y_{e_i} \in \{0, 1\}$, denoting whether $e_i$ has been killed by police or not based on the distant supervision procedure ($y_{e_i}$ is set to 1 if $e_i$ is deemed to be killed by police via distant supervision and 0 otherwise). For convenience, let $Y = \left\{y_{e_i}\right\}_{i=1}^{N}$. Our goal is to use $E$, $C$ and $Y$ as training data to generate a model that can predict whether a new entity $e$ is a victim of a police killing incident or not, given its sentence container $c$ in some corpus. In machine learning, this essentially amounts to building models to estimate the probability $P(y_e = 1 | c)$.

We will first introduce the hierarchical LSTM model for this problem, and then describe the supervised attention mechanisms with semantical word lists and dependency trees.

\begin{figure}[t]
\addtolength{\abovecaptionskip}{-4.0mm}
\addtolength{\belowcaptionskip}{-5.1mm}
\includegraphics[scale=0.6]{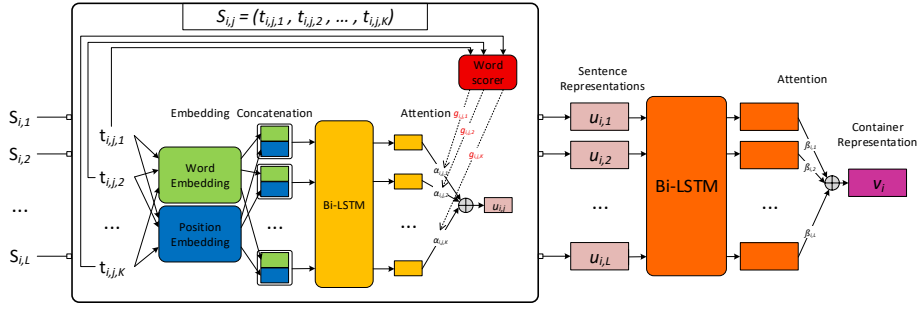}
\centering
\caption{Hierarchical LSTMs with Supervised Attention for Police Killing Detection.}
\label{fig:system-illustration}
\end{figure}

\subsection{Hierarchical LSTMs}
\label{sec:vanilla}
Each entity $e_i \in E$ along with its container $c_i \in C$ constitute an example for the model. Let $c_i = (s_{i, 1},s_{i, 2}, \ldots, s_{i, L})$ be the list of sentences in $c_i$ where $L$ is the number of sentences and $s_{i,j}$ is the $j$-th sentence in the container $c_i$. Each sentence $s_{i,j}$ is in turn a word/token sequence: $s_{i,j} = (t_{i, j, 1},t_{i, j, 2},\ldots,t_{i, j, K})$ with $K$ as the number of the tokens and $t_{i,j,k}$ as the $k$-th token in the sentence $s_{i,j}$. For each sentence $s_{i,j}$, let $k_{i,j}$ be the index of the name of the entity $e_i$ (i.e, the token $t_{i,j,k_{i,j}}$). Note that the order of the sentences $s_{i,j}$ in $c_i$ is obtained by sorting the sentences according to the download time of their corresponding documents in $D$\footnote{Such documents are retrieved via running pre-defined search queries once per hour throughout 2016 \cite{keith2017identifying}.}. This helps to partially retain information about the order of being mentioned of the entities in the sentence containers in the corpus.

The hierarchical LSTM model in this work processes one entity $e_i$ and its corresponding sentence container $c_i$ at a time. For each entity $e_i$, the operation of the model can be divided into two main components: Embedding and Attention. Figure \ref{fig:system-illustration} shows an overview of the proposed model.

\subsubsection*{Embedding}
In this component, each token $t_{i, j, k}$ of a sentence $s_{i,j}$ in $c_i$ is transformed into an embedding vector $x_{i, j, k} = [q_{i, j, k}, p_{i, j, k}]$, which is the concatenation of the word embedding vector $q_{i, j, k}$ and the position embedding vector $p_{i, j, k}$ \cite{Nguyen:15a}. These vectors are obtained as follows:

\textit{Word embedding vector}: $q_{i, j, k}$ is obtained by taking the column vector corresponding to $t_{i,j,k}$ in the pre-trained word embedding matrix $W_e$ (i.e, \texttt{word2vec} in our case): $q_{i, j, k} = W_e[t_{i, j, k}]$.

\textit{Position embedding vector}: $p_{i, j, k}$ captures the relative distance $k-k_{i,j}$ from the token $t_{i, j, k}$ to the entity name token $t_{i, j, k_{i,j}}$ in the sentence: $p_{i, j, k} = W_d[k-k_{i,j}]$, the $(k-k_{i,j})-th$ column in the position embedding matrix $W_d$ ($W_d$ is randomly initialized in this work) \cite{Nguyen:18}.

Once each token $t_{i,j,k}$ has been transformed into a vector, the corresponding input sentence $s_{i,j}$ would become a sequence of vector $(x_{i,j,1}, x_{i,j,2}, \ldots, x_{i,j,K})$. This allows us to view the container $c_i$ as an ordered list of vector sequences for its sentences $(s_{i, 1},s_{i, 2}, \ldots, s_{i, L})$. 

\subsubsection*{Attention}
The attention component processes the list of vector sequences produced in the previous step for $c_i$ at two levels (i.e, the word level and the sentence level) to produce a single representation vector for $c_i$.

The word level component consumes each vector sequence of $c_i$ once at a time to compute the representation vector for the corresponding sentence. For a vector sequence $(x_{i,j,1}, x_{i,j,2}, \ldots, x_{i,j,K})$ of $s_{i,j}$, the model architecture consists of two LSTMs \cite{Hochreiter:97} (i.e, the forward LSTM and the backward LSTM) that operate over two different directions of the vector sequence (bidirectional LSTMs) \cite{Cho:14,Nguyen:16a}. The resulting hidden vector sequences of the forward and backward LSTMs are concatenated at each position, generating the hidden vector sequence $(h_{i,j,1}, h_{i,j,2}, \ldots, h_{i,j,K})$ for the input vector sequence of $s_{i,j}$.




In order to combine the hidden vectors $h_{i,j,k}$, the attention mechanism computes the weighted sum of the hidden vectors to obtain a single representation vector for the input sentence $s_{i,j}$ \cite{Bahdanau:15}. Specifically, each hidden vector $h_{i, j, k}$ is given a weight $\alpha_{i,j,k}$ to estimate its contribution for the final representation of the container $c_i$ for the problem of police killing recognition. In this work, the weight $\alpha_{i,j,k}$ is computed by:

\begin{equation}
\label{eq:learned_weight}
\alpha_{i,j,k}=\frac{\exp(a_{i,j,k}^\top w_{a})}{\sum_{k'}{\exp(a_{i,j,k'}^\top w_{a})}}
\end{equation}
where:
\begin{equation}
\label{eq:att_rep}
a_{i,j,k}=\tanh(W_{att} h_{i,j,k} + b_{att})
\end{equation}
In such equations, $W_{att}$, $b_{att}$ and $w_{a}$ are the attention parameters at the word level that are learnt in the training process. Eventually, the representation vector $u_{i,j}$ of the sentence $s_{i,j}$ is:
\begin{equation}
\label{eq:sent_rep}
u_{i,j}=\sum_k\alpha_{i,j,k}h_{i,j,k}
\end{equation}

Once the word level component has been completed, every sentence $s_{i,j}$ in the container $c_i$ would have a corresponding representation vector $u_{i,j}$. Such sentence representation vectors $u_{i,j}$ altogether form a new sequence of vector $(u_{i, 1}, u_{i, 2}, \ldots, u_{i, L})$ to represent the container $c_i$. At the sentence level, $(u_{i, 1}, u_{i, 2}, \ldots, u_{i, L})$ is processed in the same way we process the vector sequence $(x_{i,j,1}, x_{i,j,2}, \ldots, x_{i,j,K})$ at the word level to produce the sentence representation vector $u_{i,j}$. In particular, $(u_{i, 1}, u_{i, 2}, \ldots, u_{i, L})$ would be first passed to a bidirectional LSTM model to obtain the hidden vector sequence $(h_{i, 1}, h_{i, 2}, \ldots, h_{i, L})$. This is in turn fed into the attention component to obtain the attention weights $(\beta_{i, 1}, \beta_{i, 2}, \ldots, \beta_{i, L})$ (i.e, similar to Equations \ref{eq:learned_weight} and \ref{eq:att_rep}). Finally, we compute the vector representation $v_i$ for the sentence container $c_i$ via the weighted sum: $v_i=\sum_j\beta_{i,j} h_{i,j}$. As the sentences in $c_i$ are sorted by their appearance time, the attentional bidirectional LSTMs at the sentence level attempt to estimate the contribution of each sentence $s_{i,j}$ in $c_i$ for the representation vector $v_i$ with respect to its past and future context information (i.e, the sentences before and after $s_{i,j}$ in $c_i$).

The container representation vector $v_i$ for $c_i$ allows us to compute the probability $P_i$ of $e_i$ being killed by police: $P_i = P(y_{e_i} = 1 \mid c_i) = \sigma(W_{out}v_i + b_{out})$ where $W_{out}$ and $b_{out}$ are the model parameters, and $\sigma$ is the logistic function. In order to train the hierarchical LSTM model in this section, we use the cross-entropy between the predicted labels and the golden labels as the loss function:
\begin{equation}
\label{eq:original_loss}
L_{c} = - \sum_{e_i} y_{e_i} \log (P_i) + (1-y_{e_i}) \log (1-P_i)
\end{equation}

\subsection{Supervised Attention}
\label{sec:supervised_attention}

The position embeddings $p_{i,j,k}$ in the initial representation vectors of the tokens is crucial to the hierarchical LSTM model as it helps to indicate the positions of the entity names of interests in the sentences. Technically, the position embeddings would tell the model to pay more attention to the words in the entity names by assigning higher attention weights to the representation vectors of the entity name tokens $h_{i,j,k_{i,j}}$ at the word level. Unfortunately, in the experiments, we find that this procedure might lead to extremely high weights for the tokens of the entity names, leaving essentially negligible weights for the other important context words in the sentences. The consequence is the incorrect predictions of the model for the entities in such situations. In order to solve this problem, in this work, we seek to use linguistic intuitions to obtain the rough estimations of the attention weights for the words in the sentences (intuitive attention weights), quantifying our belief about the importance of the words in the sentences for the problem of police killing recognition. The intuitive attention weights would then be used to guide the attention weights computed by the hierarchical LSTM model at the word level (i.e, Equation \ref{eq:learned_weight}), penalizing the model parameters with significant difference between the two types of attention weights.

Formally, for an entity $e_i$ with the sentence container $c_i$, suppose that we can obtain the intuitive attention weights $(g_{i,j,1}, g_{i,j,2}, \ldots, g_{i,j,K})$ for the tokens of every sentence $s_{i,j} = (t_{i, j, 1},t_{i, j, 2},\ldots,t_{i, j, K})$ in $c_i$ ($\sum_k g_{i,j,k} = 1 \forall i, j$). The difference between the intuitive attention weights and the model attention weights in Equation \ref{eq:learned_weight} for $e_i$ can be computed via the sum of the squared element difference: $L_{i} = \sum_{j,k}(g_{i,j,k} - \alpha_{i,j,k})^2$.

Our goal is to minimize this difference so that the model attention weights can encode our intuition about the importance of the words in the sentences. This essentially translates into an integrated loss function to train the hierarchical LSTM model, attempting to minimize the loss function in Equation \ref{eq:original_loss} and the attention weight difference simultaneously: $L = L_c + \lambda \sum_i L_i$ where $\lambda$ is a penalty coefficient to control the effect of the attention difference.

\subsubsection*{Generating Intuitive Attention Weights}

The previous section has described the supervised attention mechanism for the hierarchical LSTM model. It remains to investigate the methods to obtain the intuitive attention weights. An important characteristics of the intuitive attention weights is that they should assign high weights to the linguistically important words for police killing recognition. In this work, we first start by selecting the important words in the sentences based on our intuition. Afterward, non-negative scores are given to the selected words and their neighbors in the sentences, leaving zero scores for other words. These scores are then normalized to ensure a sum of 1. In order to generate the non-negative scores for the selected words and their neighbors, we employ a Gaussian distribution $\mathcal{N}(\mu, \sigma^2)$\cite{mi2016supervised,Liu:17} with the mean $\mu$ and the standard deviation $\sigma$ so that the closer neighboring words would receive higher scores.

\subsubsection*{Selecting Important Words}

This section presents two methods to select important words in the sentences of the containers for the problem of police killing detection. The first method relies on the semantical aspect while the second method concerns the syntactical heuristics. We will compare these methods in the experiments.

{\bf Semantical Aspect}: The two concepts that are most related to our problem of police killing recognition would naturally be ``\textit{police}'' and ``\textit{killing}''. It is therefore intuitive to consider these two words (``\textit{police}'' and ``\textit{killing}'') and their similar ones as the important words for our problem. Consequently, for every sentence in the containers, we search for such important words and use the matched word as the selected words. Following \cite{keith2017identifying}, we generate the lists of similar words for ``\textit{police}'' and ``\textit{killing}'' by looking up the nearest words in cosine distance via the word vectors pre-trained with \texttt{word2vec} and its corpus. Note that this method also includes the names of the entities $e_i$ in the list of selected important words due to their roles in specifying the entities of interests in the sentences. 

{\bf Syntactical Aspect}: The semantical aspect with the lists of similar words for police killing detection might be helpful for the sentences that express a police killing instance (positive sentences) and contain the similar words. However, for the negative sentences that do not mention police killing incidents, the appearance of the similar words in the lists for ``\textit{killing}'' and ``\textit{police}'' might be harmful to the supervised attention mechanisms as the emphasis on such words might lead to an incorrect impression to consider the sentence as actually positive. For instance, consider the following negative sentence with the words in the similar word lists written in bold\footnote{Throughout this work, the names of the entities of interests would be replaced by ``\textit{TARGET}'' while the other entity mentions in the sentences would be substituted by ``\textit{PERSON}'' for generalization.}:


\begin{center}
\textit{Marion \textbf{Police} Department have arrested \textbf{TARGET} , 20 , of Marion , in connection with the \textbf{fatal} hit.}
\end{center}

In this sentence, the extreme emphasis on ``\textit{Police}'', ``\textit{TARGET}'' and ``\textit{fatal}'' might lead to the incorrect prediction that this sentence is expressing a fatal event caused by police. In order to overcome this issue, we observe that police killing recognition can be seen as a relation identification problem \cite{Bunescu:05}, attempting to decide whether the entities of interests (i.e, the ``\textit{TARGET}'') has a semantical relation of ``\textit{killed by}'' with the similar words of ``\textit{police}'' (if any) in the sentence or not. In such relation identification problem, it has been shown that the shortest dependency path connecting the two word of interests (i.e, the words ``\textit{TARGET}'' and ``\textit{police}'' in our case) in the dependency trees involve the most important context words for the problem \cite{Bunescu:05}. Consequently, in this work, we propose to select the words along the shortest dependency paths between the entity name of interests (i.e, ``\textit{TARGET}'') and the similar words of ``\textit{police}'' in the sentence as the guidance words for the supervised attention mechanism for police killing recognition (dependency tress are obtained via Stanford CoreNLP \cite{schuster2016enhanced} in this work). 

In the example sentence above, the shortest dependency path between ``\textit{TARGET}'' and ``\textit{police}'' is: \textit{Police $\rightarrow$ Department $\rightarrow$ arrested $\rightarrow$ TARGET}. It is clear in this situation that the words along the path do not suggest a police-caused killing (i.e, not a ``\textit{killed by}'' relation between ``\textit{TARGET}'' and ``\textit{Police}''). Consequently, the attention of the models to such words would lead to a correct prediction in this case.

On the other hand, for the positive sentences, it might be the case that the words along the dependency paths help to include some words that are crucial to police killing prediction, but do not appear in the semantical word lists.

\subsubsection*{Baselines}

For experimental purposes, we call the hierarchical LSTM model without supervised attention as ``\textit{H-LSTM}''. The hierarchical LSTM model with the supervised attention mechanism would then be called ``\textit{H-LSTM+SemAtt}'' and ``\textit{H-LSTM+SynAtt}'' depending on whether the semantical aspect (i.e, the lists of similar words) or the syntactical aspect (i.e, the dependency paths) is used to obtain the important words. In order to further demonstrate the benefits of supervised attention for police killing detection, we consider two more baseline models when the model attentions at the word level are excluded from ``\textit{H-LSTM+SemAtt}'' and ``\textit{H-LSTM+SynAtt}'', resulting in the models  ``\textit{Mean+SemAtt}'' and ``\textit{Mean+SynAtt}''. In particular, in such models, the sentence representation vector $u_{i,j}$ for the sentence $s_{i,j}$ would not be obtained via the attention-based weighted sum in Equation \ref{eq:sent_rep}. In contrast, $u_{i,j}$ would be computed using the mean vector of the vector set $\{h_{i,j,k_1}, h_{i,j,k_2}, \ldots, h_{i,j,k_I}\}$ from LSTMs where $k_1, k_2, \ldots, k_I$ are the indexes of the selected important words using the semantical or syntactical aspect for the sentence $s_{i,j} = (t_{i, j, 1},t_{i, j, 2},\ldots,t_{i, j, K})$: $u_{i,j} = 1/I \sum_{m=1}^{I} h_{i,j,k_m}$.
\subsection{Training}

\label{sec:training}
We train all the models in this work using stochastic gradient descent with shuffled mini-batches, the Adam update rule, back-propagation and dropout. Non-embedding weights are also imposed to gradient clipping to rescale their $l2$-norms if they exceed a predefined threshold.

\section{Experiments}
\label{sec:exp}
\subsection{Datasets and Parameters}
\label{sec:data}
We evaluate the models in this paper using the police killing dataset released by \cite{keith2017identifying}.

As there is no development data for this dataset, we divide the original training data into two parts, for which one part is used for training data while the other part functions as the development data. We use these newly-generated training data and development data to select the parameters for the models. For the comparison with the state-of-the-art models in \cite{keith2017identifying}, we utilize the original training data and test data with the chosen parameters from the development experiments to ensure a compatible comparison. We use the same procedure to split the original training data for development as those employed by \cite{keith2017identifying} to generate the original dataset. In particular, we first sort all the positive entities in the original training data using the descending order of their death times. Afterward, we identify the death time of the entity at the bottom of the top 20\% in the sorted list. The date we found is used as the split point. All the entities with the download time or death time after this date are utilized as the development data.


The parameters we found in the development experiments are as follow. The dimensionality parameters include: 8 dimensions for position embedding vectors, 300 dimensions for word embedding vectors, 256 hidden units for the LSTMs and 64 dimensions for the attention vectors. For supervised attention, the penalty coefficient $\lambda$ is set to 1.0 while the neighbor window $T$ for generating intuitive attention weights is 2. The threshold for gradient clipping is set to 5.0 while the dropout rate is 0.5. The mean and standard deviation of the Gaussian distribution have the values of $\mu=0$ and $\sigma=1.0,$ respectively.

\subsection{Evaluating the Models}
\label{sec:architecture}

We evaluate the models (i.e, \textit{H-LSTM}, \textit{H-LSTM+SemAtt}, \textit{H-LSTM+SynAtt}, \textit{Mean+SemAtt} and \textit{Mean+SynAtt}) using the generated development data. Table \ref{tab:architecture_evaluation} reports the performance. 

\begin{table}[h]
\addtolength{\abovecaptionskip}{-2.0mm}
\addtolength{\belowcaptionskip}{-10mm}
\centering
\begin{tabular}{lccc}
\Xhline{2\arrayrulewidth}
Models          & Precision & Recall & F1    \\ \Xhline{2\arrayrulewidth}
H-LSTM+SynAtt & 0.497     & 0.381  & \textbf{0.431} \\
H-LSTM+SemAtt & 0.366     & 0.504  & 0.424 \\
H-LSTM         & 0.460     & 0.377  & 0.414 \\
Mean+SynAtt    & 0.428     & 0.372  & 0.398 \\
Mean+SemAtt    & 0.346     & 0.399  & 0.371 \\ \Xhline{2\arrayrulewidth}
\end{tabular}
\caption{\small Performance of the models on the development data. The comparisons in this table are significant with $p < 0.05$.}
\label{tab:architecture_evaluation}
\end{table}

\begin{table}[b]
\addtolength{\abovecaptionskip}{-2.0mm}
\addtolength{\belowcaptionskip}{-7.0mm}
\centering
\small
\begin{tabular}{lcccc}
\Xhline{2\arrayrulewidth}
Models                          & Precision      & Recall         & F1             & AUC            \\ \Xhline{2\arrayrulewidth}
\textbf{H-LSTM+SynAtt} & 0.442 & 0.288 & \textbf{0.349} & \textbf{0.211} \\ 
H-LSTM+SemAtt           & 0.342          & 0.325          & 0.333          & 0.199          \\ 
H-LSTM                        & 0.419          & 0.259          & 0.320          & 0.194          \\ \Xhline{2\arrayrulewidth}
soft-LR (EM)                   & 0.447          & 0.243          & 0.316          & 0.193          \\ 
soft-CNN (EM)                  & 0.268          & 0.265          & 0.267          & 0.164          \\ \Xhline{2\arrayrulewidth}
\end{tabular}
\caption{\small Performance comparison on test data. AUC is the area under the precision/recall curve. The comparison in this table is significant with $p < 0.05$.}
\label{tab:performance_comparison}
\end{table}

There are three major observations from the table. First, the performance of the baseline models \textit{Mean+SynAtt} and \textit{Mean+SemAtt} are much worse than the other models, demonstrating that the attentions in Equation \ref{eq:learned_weight} are crucial to problem of police killing detection. Second, both \textit{H-LSTM+SemAtt} and \textit{H-LSTM+SynAtt} are better than \textit{H-LSTM} (improvements of 1\% and 1.7\% on the absolute F1 scores for \textit{H-LSTM+SemAtt} and \textit{H-LSTM+SynAtt} respectively). This shows the benefits of the proposed supervised attention mechanisms with semantical and syntactical guidance in this work. Third, we see that \textit{H-LSTM+SynAtt} outperforms \textit{H-LSTM+SemAtt}, suggesting that the syntactical guidance with dependency trees are more effective than the semantical guidance with the word lists for supervised attention in our task. We also see that the recall of \textit{H-LSTM+SemAtt} is much better than that of \textit{H-LSTM+SynAtt}. We attribute this phenomenon to the fact that the dataset in \cite{keith2017identifying} involves many sentences with the words in the similar words lists for ``\textit{police}'' and ``\textit{killing}''. This biases the supervised attention in \textit{H-LSTM+SemAtt} to associate the appearance of such words with the positive entities and leads to the high recall for this method. Due to the poor performance of \textit{Mean+SynAtt} and \textit{Mean+SemAtt} in this development experiment, we only consider the other models (i.e, \textit{H-LSTM}, \textit{H-LSTM+SemAtt} and \textit{H-LSTM+SynAtt}) in the following experiments.

\subsection{Comparing to the State of the Art}
\label{sec:compare}

This section compares our proposed models with the state-of-the-art models for police killing recognition. Such state-of-the-art models include \textit{soft-RL} and \textit{soft-CNN} that both apply the Expectation Maximization algorithm, but employ logistic regression and convolutional neural networks (respectively) for sentence classifiers \cite{keith2017identifying}. Table \ref{tab:performance_comparison} shows the performance the models. Note that the performance in this section is obtained using the original training data and test data in \cite{keith2017identifying}.

As we can see from the table, the conclusions we have for the models \textit{H-LSTM}, \textit{H-LSTM+SemAtt} and \textit{H-LSTM+SynAtt} in the previous section still hold in this case on the test data, thus further confirming those observations for police killing recognition. We also see that although \textit{H-LSTM} does not use supervised attention, its performance is comparable with the best model \textit{soft-LR} in \cite{keith2017identifying}. This is significant as \textit{H-LSTM} does not employ any hand-crafted features while \textit{soft-LR} needs to resort to complicated hand-designed features to perform well. The best performance is achieved with the \textit{H-LSTM+SynAtt} model with an improvement of 3.3\% in the absolute F1 measure over the best model \textit{soft-LR} in \cite{keith2017identifying}. This testifies to the effectiveness of our proposed model in this work, featuring hierarchical LSTMs, supervised attention and syntactical guidance.


\subsection{Analysis}
\label{sec:analysis}

In order to demonstrate the effectiveness of supervised attention and syntactical guidance for police killing detection, this section visualizes the attention weights $\alpha_{i,j,k}$ in Equation \ref{eq:learned_weight} for the words in several sentences in the test data.

\subsubsection*{The Effect of Supervised Attention}

\begin{figure}[h]
\addtolength{\abovecaptionskip}{-4.0mm}
\addtolength{\belowcaptionskip}{-10.mm}
\includegraphics[scale=0.075]{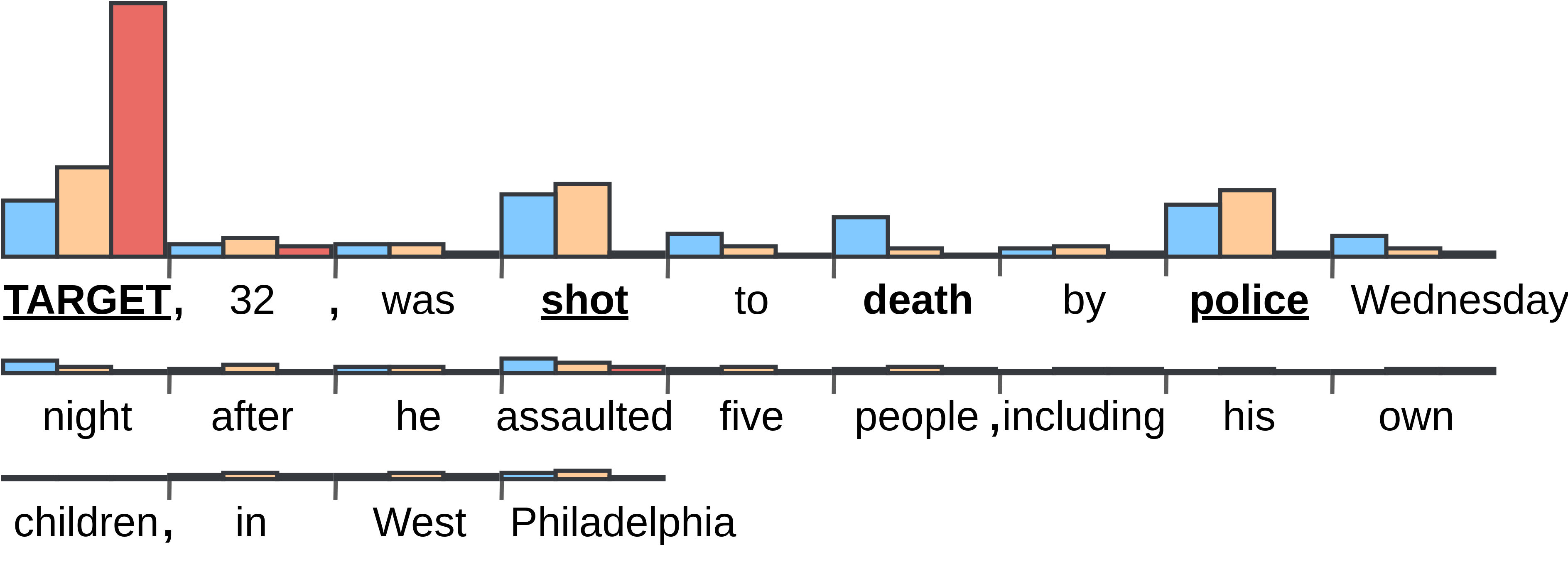}
\centering
\caption{\small Attention weight visualization. The underlined and bold words are important words selected by \textit{H-LSTM+SemAtt} and \textit{H-LSTM+SynAtt} respectively. The  blue, orange and red columns represents the word attention weights computed by \textit{H-LSTM+SynAtt}, \textit{H-LSTM+SemAtt}, and \textit{H-LSTM} respectively. Weights of punctuations  are not shown.}
\label{fig:analysis-1}
\end{figure}

Figure \ref{fig:analysis-1} indicates the attention weights computed by the models \textit{H-LSTM}, \textit{H-LSTM+SemAtt} and \textit{H-LSTM+SynAtt} for the words in an example sentence. This sentence corresponds to an entity in the test set that is correctly predicted (as being killed by police) by \textit{H-LSTM+SynAtt}, but is incorrectly predicted by \textit{H-LSTM} and \textit{H-LSTM+SemAtt}. As we can see from the figure, \textit{H-LSTM} fails in this case as it reserves a very high weight for the ``\textit{TARGET}'' and essentially ignores the other words. This phenomenon is quite popular for \textit{H-LSTM} and demonstrates the needs for supervised attention mechanisms as being motivated in the previous sections. In addition, \textit{H-LSTM+SemAtt} cannot use this sentence as an evidence to make a correct prediction in this case as it mainly attends to the words in the similar word lists (i.e,``\textit{TARGET}'', ``\textit{shot}'' and ``\textit{police}'') and misses the word ``\textit{death}''. This is undesirable as ``\textit{death}'' is the only clue showing that the victim of the shooting in this sentence is actually dead. \textit{H-LSTM+SynAtt} is successful in this case as it is able to assign high weights to such important words along the dependency paths. This demonstrates our arguments in Section 3, showing the benefits of \textit{H-LSTM+SynAtt} to suggest important words that cannot be captured by \textit{H-LSTM+SemAtt} for police killing recognition.

\subsubsection*{Semantical vs. Syntactical Guidance}

\begin{figure}[h]
\addtolength{\abovecaptionskip}{-4.0mm}
\addtolength{\belowcaptionskip}{-6.0mm}
\includegraphics[scale=0.075]{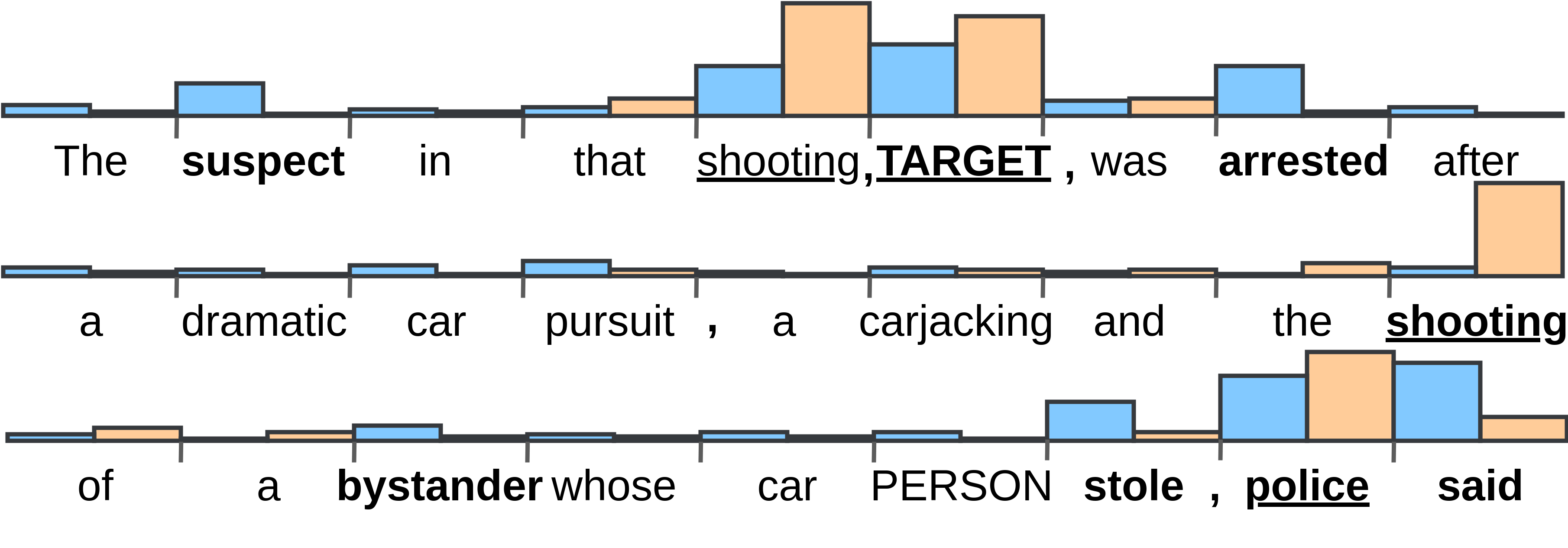}
\centering
\caption{\small Attention weight visualization. The conventions in Figure \ref{fig:analysis-1} do apply here.}
\label{fig:analysis-2}
\end{figure}

The previous part has shown the advantages of \textit{H-LSTM+SynAtt} over \textit{H-LSTM+SemAtt} for positive entities. This section focuses on the benefits of \textit{H-LSTM+SynAtt} for the negative entities. Figure \ref{fig:analysis-2} illustrates the attention weights that \textit{H-LSTM+SemAtt} and \textit{H-LSTM+SynAtt} assign to the words of an example sentence in the test data.

The entity of this sentence is negative that has been correctly recognized by \textit{H-LSTM+SynAtt}, but has been incorrectly predicted by \textit{H-LSTM+SemAtt}. As suggested in the figure, the failure of \textit{H-LSTM+SemAtt} is due to its very high weights on ``\textit{shooting}'', ``\textit{TARGET}'' and ``\textit{police}'', ignoring the effect of the words ``\textit{said}'' and ``\textit{arrested}'' that clearly negate the involvement of police in this shooting. \textit{H-LSTM+SynAtt} can attend to such important words as they belong to the dependency paths between ``\textit{police}'' and ``\textit{TARGET}'' in this case.

\section{Conclusions}
\label{sec:conclusions}
We propose a novel deep learning model for the problem of police killing recognition. The proposed model involves hierarchical LSTMs to model the multiple sentences in the sentence containers of the entities. We introduce novel supervised attention mechanisms based on semantical and syntactical aspects for this problem. The experimental results demonstrate the effectiveness of the proposed models and lead to the state-of-the-art performance for police killing detection. In the future, we plan to apply the proposed method in a real system and extend it to other types of events (e.g, protests, epidemics).

\bibliographystyle{acl}
\bibliography{coling2018}








\end{document}